\documentclass[a4paper,conference]{IEEEtran}
\ifCLASSINFOpdf
  \usepackage[pdftex]{graphicx}
\else
\fi
\usepackage{amsmath}
\usepackage{amssymb}
\usepackage{algorithmic}
\usepackage{algorithm}
\ifCLASSOPTIONcompsoc
  \usepackage[caption=false,font=normalsize,labelfont=sf,textfont=sf]{subfig}
\else
  \usepackage[caption=false,font=footnotesize]{subfig}
\fi

\usepackage{multirow}
\usepackage{booktabs}

\begin{document}
%
\title{KRNet: Towards Efficient Knowledge Replay}



%
\author{\IEEEauthorblockN{Yingying Zhang, Qiaoyong Zhong\textsuperscript{$\star$}, Di Xie and Shiliang Pu}
\IEEEauthorblockA{Hikvision Research Institute, Hangzhou, China\\
\{zhangyingying7,zhongqiaoyong,xiedi,pushiliang.hri\}@hikvision.com}}


\maketitle

{\let\thefootnote\relax\footnotetext{$^\star$Corresponding author.}}

\begin{abstract}
  The knowledge replay technique has been widely used in many tasks such as continual learning and continuous domain adaptation. The key lies in how to effectively encode the knowledge extracted from previous data and replay them during current training procedure. A simple yet effective model to achieve knowledge replay is autoencoder. However, the number of stored latent codes in autoencoder increases linearly with the scale of data and the trained encoder is redundant for the replaying stage. In this paper, we propose a novel and efficient knowledge recording network (KRNet) which directly maps an arbitrary sample identity number to the corresponding datum. Compared with autoencoder, our KRNet requires significantly ($400\times$) less storage cost for the latent codes and can be trained without the encoder sub-network. Extensive experiments validate the efficiency of KRNet, and as a showcase, it is successfully applied in the task of continual learning.
\end{abstract}


%
\IEEEpeerreviewmaketitle

\section{Introduction}
	\label{sect:intro}
    In recent years, deep neural networks (DNNs) have achieved great success in various multi-class classification tasks. However, most current models can only solve the classification tasks with fixed categories and data distribution because of the serious problem of catastrophic forgetting~\cite{goodfellow2014an,mccloskey1989catastrophic}. In practice, the learning environment is complicated and changeable, which requires the model to be able to deal with different learning tasks continually. From one perspective, catastrophic forgetting is mainly caused by the discriminative nature of a classifier, which learns to distinguish objects of different categories rather than to remember them. Once the network is forced to learn new discernibility, the previous ones will be overridden. Therefore, replaying the information (knowledge) of previous data is one of the most effective methods for continual learning. The main idea is to learn a model which can generate images \cite{shin2017continual,wu2018memory} or features \cite{liu2020generative} of the previous task data, and mix them up with the new task data when training the current model. For data replay, both generative adversarial network (GAN)~\cite{goodfellow2020generative} and autoencoder (AE)~\cite{rumelhart1988learning,hinton2006reducing} have been exploited. However, we notice the weaknesses of the two models. For GAN, it is difficult to model the distribution of samples accurately either in the image space or feature space, especially when the number of training samples is limited. The inevitable gap between generated fake samples and real samples leads to sub-optimal performance. For AE, it needs to train an extra encoder which is useless in the inference stage, and store lots of latent codes which increase linearly with the scale of data. 
	
	\begin{figure}[!t]
        \centering
		\includegraphics[width=\linewidth]{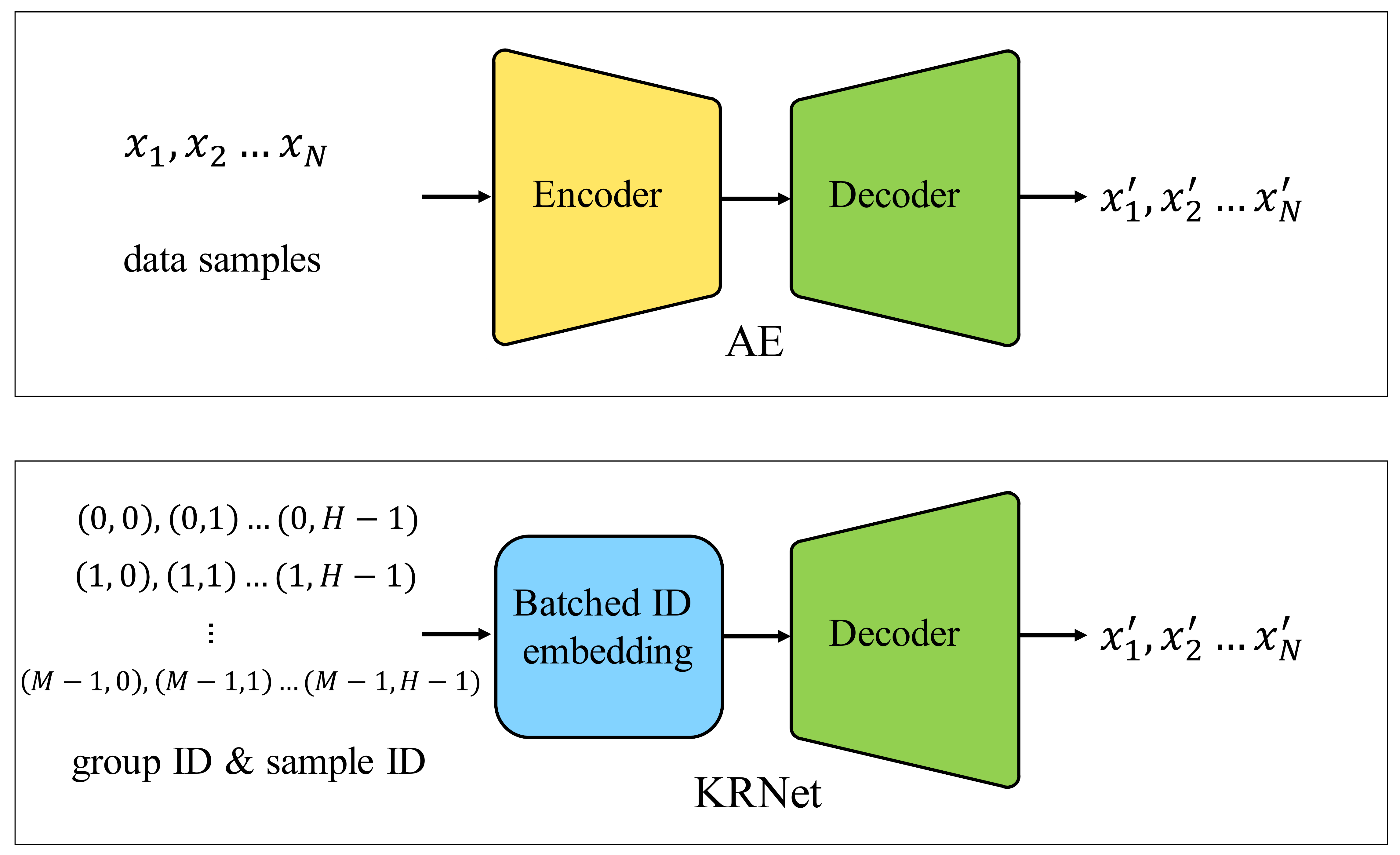}
        \caption{Difference between AE and KRNet. The encoder of AE takes each data sample as input and encodes it to a latent code. Then, the decoder maps the latent codes back to the corresponding data samples. In our KRNet, we split $N$ samples into $M$ groups (with a group size of $H$). Then the group ID and sample ID are mapped directly into an embedding vector which plays the same role as the latent code of AE by the batched ID embedding module.}
		\label{krnet_vs_ae}
	\end{figure}
	
    In this work, we propose a novel knowledge recording network (KRNet) to recite the knowledge of previous tasks. Here \emph{knowledge} typically refers to the feature maps of an intermediate layer within a deep neural network. Knowledge replay has been widely used in applications such as continual learning~\cite{liu2020generative} and continuous domain adaptation~\cite{lao2020continuous}. How to encode the knowledge efficiently without sacrificing performance is of great importance for such applications. As shown in Fig.~\ref{krnet_vs_ae}, our KRNet directly maps an arbitrary sample identity (ID) number to the corresponding feature map. By sharing the embedding vector within each group, the storage cost of the latent codes can be reduced dramatically. As discussed in Section~\ref{sec:results}, compared with AE, we are able to reduce the size of latent codes by over $400\times$. Notably, it easily enjoys the benefit of the powerful fitting capability of deep neural networks. And an overfitting on reconstruction of the input features is preferred, which is unusual in most machine learning applications. Similarly, \cite{dupont2021coin} also used this attractive nature to map pixel locations to RGB values for image compression. The main difference is that their network is a simple multilayer perceptron and a separate network is trained for each sample.
	
\begin{figure}[!t]
        \centering
		\includegraphics[width=\linewidth]{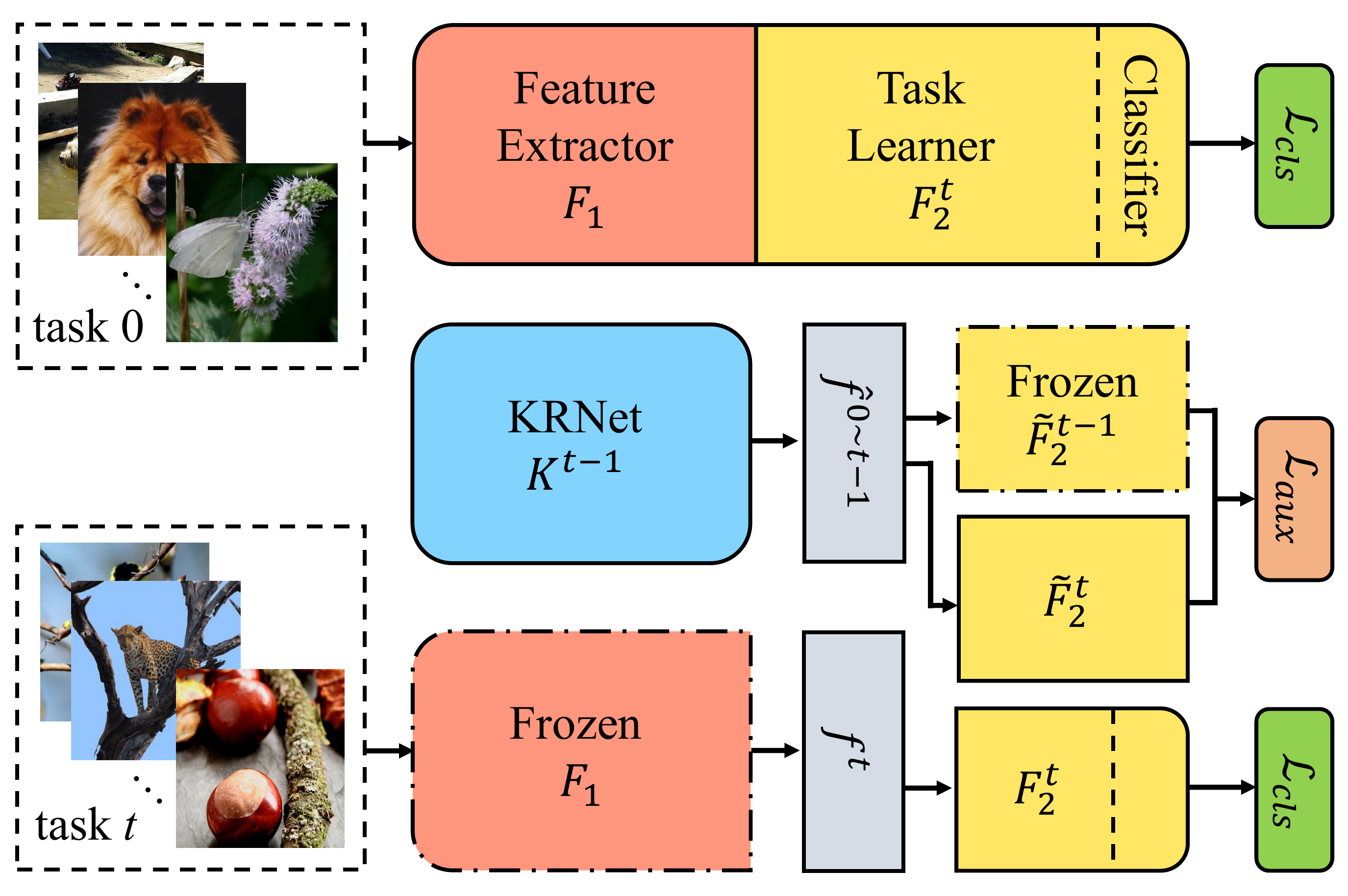}
		\caption{Pipeline of the proposed KRIL framework. KRNet is incorporated to accurately record the features of previous tasks $0\sim t-1$ and replay them in current task $t$ for incremental training.}
		\label{menet_inc_pipline}
	\end{figure}

	We also introduce an exemplary application of our KRNet in continual learning. Based on KRNet, we build a knowledge recitation-based incremental learning (KRIL) framework. As shown in Fig.~\ref{menet_inc_pipline}, we partition the entire network $F$ into a fixed feature extractor $F_1$ and an updatable task learner $F_2^t$. The features extracted with $F_1$ are accurately recorded and replayed using KRNet within a cycle of incremental learning. By direct recitation of the knowledge, the gap between the fake and real features can be reduced effectively, leading to superior performance.

	Our main contributions can be summarized as follows:
	\begin{itemize}
      \item We propose a novel knowledge recording network to encode feature maps. It achieves an extremely high compression ratio without sacrificing performance.
      \item By learning a direct mapping from an arbitrary sample ID to the corresponding datum, our KRNet is superior over autoencoder with a simpler architecture and significantly less storage cost.
		\item We showcase a successful application of KRNet in continual learning. Specifically, we introduce a new incremental learning strategy to replay the knowledge of previous tasks, which prevents catastrophic forgetting effectively.
		\item Extensive experiments on two commonly used incremental learning benchmarks validate the effectiveness of the proposed method.
	\end{itemize}

    \begin{figure*}[!t]
        \centering
		\includegraphics[width=0.85\linewidth]{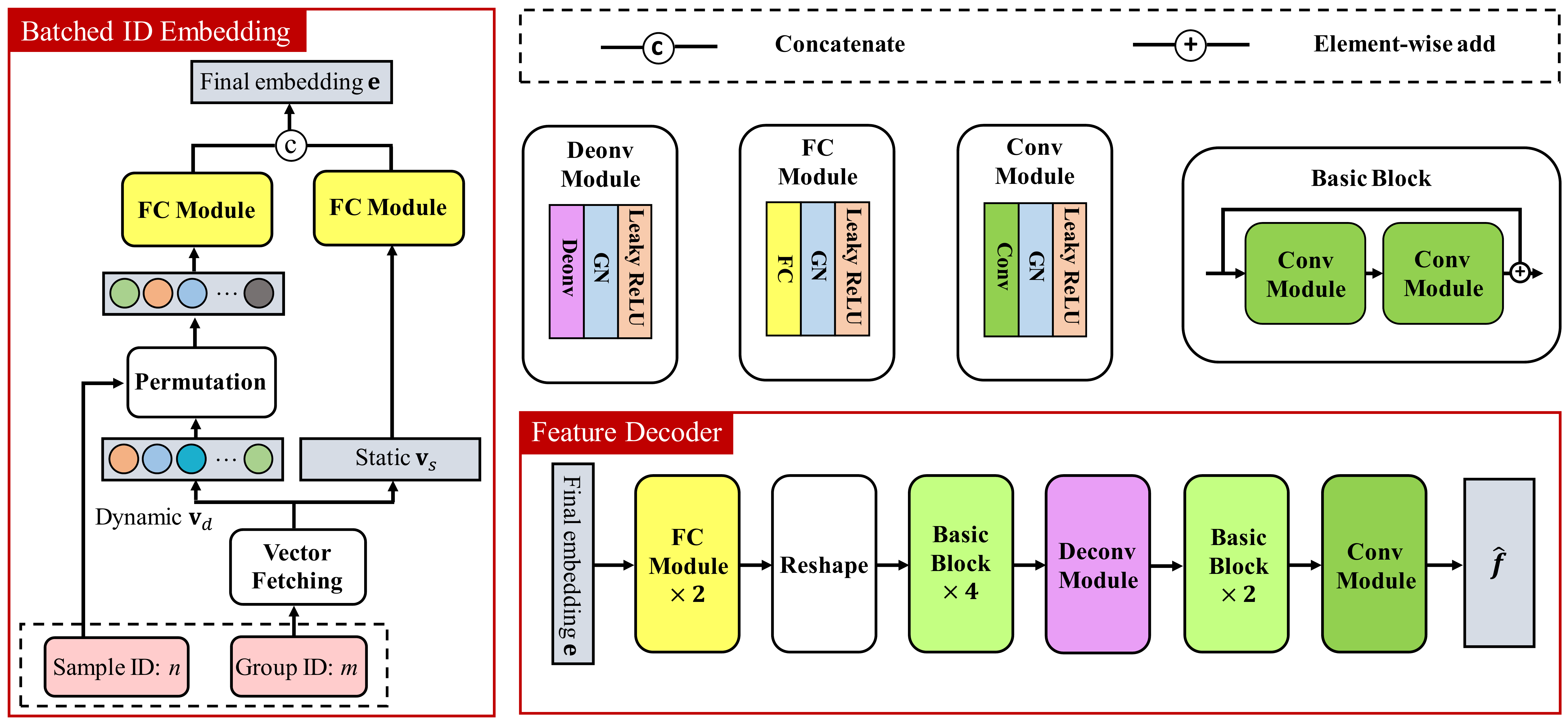}
		\caption{Architecture of KRNet. We divide all the features to be recited into groups, and assign a unique local sample ID $n$ and group ID $m$ to each feature in each group. The batched ID embedding module expands the local sample ID and group ID into the final embedding vector $\mathbf{e}$. Then the feature decoder maps the ID embedding $\mathbf{e}$ to the final features.}
		\label{menet}
	\end{figure*}
	
	\section{Related Work}
	\paragraph{Autoencoder and Generative Models}
	Autoencoder is a network designed to learn an identity function in an unsupervised way to reconstruct the input data. The model consists of an encoder and a decoder. The encoder maps the original high-dimensional input into a low-dimensional latent code, and the decoder recovers the original datum from the latent code. It is able to compress the data by reducing the data dimension for a more efficient representation. Thus, it is widely utilized in many knowledge replay scenarios. Based on AE, many variants have been developed, such as sparse AE~\cite{makhzani2014ksparse}, contractive AE~\cite{rifai2011higher} and variational AE (VAE)~\cite{kingma2014auto}. In recent years, generative adversarial network (GAN)~\cite{goodfellow2020generative} and its variants have emerged as powerful generative models. GAN is composed of a generator network and a discriminator network. Given training data drawn from a real distribution, a GAN model aims at optimizing the generator to output samples close to the real distribution.
	
	\paragraph{Knowledge Replay-based Continual Learning} 
	These methods are attracting increasing interests in the continual learning field due to their simple design and good performance. More specifically, they can be further divided into rehearsal-based methods and pseudo-rehearsal based methods. For rehearsal-based methods \cite{rebuffi2017icarl,castro2018end}, a small number of exemplars from the training samples are stored and replayed during continual training. Normally the distillation loss is employed to retain the previous knowledge and the cross entropy loss is used to learn the new classes. GEM~\cite{lopez-paz2017gradient} not only stores a subset of the observed examples from current task but also preserves the gradients of previous tasks. \cite{hou2019learning} pointed out that a crucial cause to the forgetting problem of continual learning is the imbalance between the previous and new data. They proposed a few techniques like cosine normalization to overcome this issue. \cite{belouadah2019il2m} introduced a bias correction layer to correct the output of the original fully-connected layer to address the data imbalance.
	
	In contrast to rehearsal-based methods, pseudo-rehearsal based methods do not make use of any real images. Instead, they attempt to solve the forgetting problem by generating fake data which share similar distribution to real data using generative models. \cite{shin2017continual} proposed an image replay method that uses GAN and an auxiliary classifier to generate samples belonging to previous classes. In order to generate images more suitable for incremental learning, \cite{wu2018memory} incorporated a conditional GAN. There are also approaches which synthesize the intermediate features rather than images. FearNet~\cite{kemker2017fearnet} uses a VAE to generate features of previous tasks for memory replay. The limitation of this approach is the strong assumption of a Gaussian distribution of the data and the need of pre-trained models. \cite{liu2020generative} proposed a more general feature replay-based method which employs a GAN to generate the features before the classifier. Feature replay-based methods are the most relevant to the proposed KRIL framework, but the replay mode and incremental training procedure are different. Our KRIL approach directly recites the features of a shallow layer rather than modeling the distribution of high-level semantic features of the final layers in the network. By taking advantage of the task-agnostic characteristic of low-level features, KRIL circumvents the knowledge distillation of the backbone network used in \cite{liu2020generative}.

	\section{Our Method}
	\label{mehtod}
	In this section, we first elaborate the detailed architecture of the proposed KRNet. Then we introduce its application in the class-incremental learning setting. 
	\subsection{KRNet}
	To replay knowledge of previous data, a straightforward choice would be the autoencoder network. Compared with autoencoder, the proposed KRNet is more compact and efficient in terms of model size and latent code size. In particular, KRNet omits the encoder and maps an arbitrary sample ID to the corresponding feature map directly. It is composed of two components, i.e. a batched ID embedding module and a feature decoder. The batched ID embedding module learns a low-dimensional embedding for each sample ID, and the feature decoder decodes the embedding into the original 3D feature map. The architecture of KRNet is illustrated in Fig.~\ref{menet} and explained in the following sections.
	
	\subsubsection{Batched ID Embedding}
	\label{SVG}
	It is challenging to learn a mapping from a low-dimensional space to a high-dimensional space, especially in our case, where the input is a scalar. To efficiently expand the scalar into a vector, we propose a novel batched embedding scheme.
	Consider the data needing to be replayed with $N = \sum_{c=1}^{C} N^{(c)}$ samples, where $C$ is the total number of classes and $N^{(c)}$ is the number of samples of class $c$. We divide them into groups of $H$ samples at most. When the class labels of the samples are known, we make each group contain samples of the same class. The number of groups is calculated and accumulated by $M = \sum_{c=1}^{C}\lceil N^{(c)} / H\rceil $. We assign a unique group ID $m$ to each group ($0\leq m < M$). Similarly, samples in each group also have a unique local sample ID $n$ ($0\leq n < H $). For each group, we design two vectors, i.e. the static vector $\mathbf{v}_s$ and the dynamic vector $\mathbf{v}_d$. For simplicity, we set the dimension of $\mathbf{v}_s$ and $\mathbf{v}_d$ to the group size $H$ ($\mathbf{v}_s \in \mathbb{R}^H, \mathbf{v}_d \in \mathbb{R}^H$). $\mathbf{v}_s$ acts as an embedding of the group ID $m$ and is shared by all samples in the group. $\mathbf{v}_d$ acts as an embedding of the local sample ID $n$. Notably, we do not learn different $\mathbf{v}_d$ for different samples. Instead, we obtain a unique embedding for each sample by applying a permutation to a shared $\mathbf{v}_d$. The permutation can be efficiently implemented by multiplying the shared $\mathbf{v}_d$ with a permutation matrix $\mathbf{A}_n \in \mathbb{R}^{H\times H}$ associated with the local sample ID $n$. The element $a_{ij}, 0 \le i,j < H $ in $\mathbf{A}_n$ can be defined as:
	\begin{equation}
	\begin{aligned}
	a_{ij} = 
	\begin{cases}
	1,& \mathrm{if}\quad j-i = n\\
	1,& \mathrm{if}\quad i-j = H - n\\
	0,& \mathrm otherwise
	\end{cases}
	\end{aligned}
	\end{equation}
	The embeddings of the group ID and local sample ID are fed into a fully-connected (FC) module and subsequently concatenated into the final embedding $\mathbf{e} \in \mathbb{R}^{2H}$. The FC module is composed of a fully-connected layer, a group normalization (GN)~\cite{wu2018group} layer and a Leaky-ReLU~\cite{Maas13rectifiernonlinearities} layer. It is worth noting that we only need to store $M$ static and dynamic vectors, resulting in a total of $2M$ vectors of dimension $H$. Considering the requirement of one vector per sample in the case of autoencoder, the size of latent code is reduced dramatically.
	
	\subsubsection{Feature Decoder}
	The feature decoder maps the ID embedding $\mathbf{e}$ to the features $f$ to be recited. The detailed architecture is shown in Fig.~\ref{menet}. We first use two consecutive FC modules to project $\mathbf{e}$ to a new vector with dimension $c_0 \cdot h_0 \cdot w_0$. This vector is reshaped to an initial feature map $\hat{f}_{init}\in \mathbb{R}^{c_0 \times h_0 \times w_0}$. Then,  $\hat{f}_{init}$ is fed into four basic blocks and a deconvolution module successively to generate a feature map $\hat{f}_{mid}\in \mathbb{R}^{c \times h \times w}$ whose shape is identical to the target feature map $f$. Here, the structure of the basic block is illustrated on the top right of Fig.~\ref{menet}, and the deconvolution module consists of a deconvolution layer, a GN layer and a Leaky-ReLU layer. By applying two more basic blocks and a convolution module, we obtain the final feature map $\hat{f}$.
	
	\subsection{KRIL}
	In this section, we showcase an application of KRNet in continual learning. We call this method knowledge recitation-based incremental learning (KRIL). In a typical class-incremental learning setting, we are given a set (say $T+1$) of classification tasks, whose categories are mutually disjoint. Then a classification model is trained sequentially on the tasks. When trained on task $t$, the recognition ability of the model on all previous tasks $0, \dots, t-1$ should be maintained as much as possible. We denote the dataset of task $t$ as $D_t = \{(x_i, y_i)\}^{N_t}_{i=1}, t\in\{0, 1, \dots, T\}$. Here $x_i \in \mathcal{X} $ stands for the $i$-th image in the dataset, $y_i$ is the corresponding label and $N_t$ is the dataset size of task $t$.
	
	The pipeline of the proposed incremental learning framework is shown in Fig.~\ref{menet_inc_pipline}. As discussed in Section~\ref{sect:intro}, we partition the classification network into two sub-networks, i.e. the feature extractor $F_1$ and the task learner $F_2^t$. $F_1$ is trained once on task 0 and then fixed in the subsequent incremental tasks. $F_2^t$ is updated to learn each task. In task $t$, the model can be formulated as $\hat{y} = F_2^{t}(F_1(x)), x\in D_t$. We mingle the features of current task $f^t = \{F_1(x), x\in D_t\}$ with the features of all previous tasks recited by KRNet $ \hat{f}^{0\sim t-1}$. The combined features $f = \{f^t, \hat{f}^{0\sim t-1}\}$ are fed to $F_2^{t}$ to obtain the predictions $\hat{y} = F_2^t(f)$. Then, the cross entropy loss between the ground-truth labels $y$ and predictions $\hat{y}$ is applied to train the classifier. To improve the robustness against deviation of feature recitation, we impose an auxiliary loss to the features before the classifier. Specifically, it minimizes the mean squared error between the features predicted by current model $\tilde{F}_2^{t}$ and last model $\tilde{F}_2^{t-1}$. The total incremental learning loss is written as:
	\begin{equation}
	\label{cls-loss}
	\begin{aligned}
	\mathcal{L}_{inc} & = \mathcal{L}_{cls} + \lambda \mathcal{L}_{aux} \\
	\mathcal{L}_{cls} & = -\frac{1}{B}\sum_{i=1}^{B} y_i \log\hat{y_i} \\
	\mathcal{L}_{aux} & = \frac{1}{B}\sum_{i=1}^{B}\|\tilde{F}_2^{t-1}(\hat{f}_i)-\tilde{F}_2^{t}(\hat{f}_i)\|_2^2
	\end{aligned}
	\end{equation}
	Here, $\tilde{F}_2$ stands for the feature extraction part of $F_2$ without the classifier, $B$ denotes the batch size used during training, and $\lambda$ is the loss weight. Note that last model $\tilde{F}_2^{t-1}$ is frozen and not learnable when training current task $t$, and $\mathcal{L}_{aux}$ is applied to the recited samples $\hat{f}^{0\sim t-1}$ only.
	
	\subsection{Training Procedure of KRNet in KRIL}
	We denote the KRNet used in task $t$ as $K^{t-1}$. When training of task $t$ finishes, we need to learn an updated $K^t$ to be used in the next task $t+1$. We introduce a self-taught training strategy to absorb the new knowledge of task $t$ without forgetting the previous knowledge. That is, we use the replayed features from $K^{t-1}$ as the regression target due to inaccessibility of real samples of previous tasks. We impose two mean squared error losses to the features at different levels. The loss function can be written as:
	\begin{equation}
	\label{menet-loss}
	\begin{aligned}
	\mathcal{L}_{kr} & = \mathcal{L}_{kr1} + \gamma \mathcal{L}_{kr2} \\
	\mathcal{L}_{kr1} & = \frac{1}{B}\sum_{i=1}^{B}\|\hat{f}_i-f_i\|_2^2\\
	\mathcal{L}_{kr2} & = \frac{1}{B}\sum_{i=1}^{B}\|\tilde{F}_2^{t}(\hat{f}_i)-\tilde{F}_2^{t}(f_i)\|_2^2
	\end{aligned}
	\end{equation}
	where $f$ denotes the target features: $f = \{f^t, \hat{f}^{0\sim t-1}\}$, $\hat{f}$ stands for the predicted features by $K^t$, and $\tilde{F}_2^{t}$ is the feature extraction part of $F_2^t$ without the classifier. Note that unlike Eq.~\eqref{cls-loss}, training of $\tilde{F}_2^{t}$ has finished, and the parameters in it are frozen when training $K^t$. 
	The overall training procedure of the proposed incremental learning framework is described in Algorithm~\ref{alg-menet}.

	\begin{algorithm}[t]
		\caption{Knowledge recitation for incremental learning.}
		\label{alg-menet}
		\textbf{Input}: Dataset sequence $D_t = \{(x_i, y_i)\}^{N_t}_{i=1}$, $t=0, \dots, T$\\
		\textbf{Output}: Classification network $F$ partitioned into $F_1$ and $F_2$, KRNet $K$
		\begin{algorithmic}[1]
			\STATE Train the base classification network $F^0 (F_1$ and $F_2^0)$ on $D_0$ using the cross entropy loss.
			\STATE Train the base KRNet $K^0$ on the features extracted by $F_1$ using the loss in Eq.~\eqref{menet-loss}.
			\FOR { $t=1, \dotsc, T$}
			\STATE Replay previous features $\hat{f}^{0\sim t-1}$ using $K^{t-1}$.
			\STATE Extract features of current task $f^{t}$ using $F_1$.
			\STATE Train $F_2^{t}$ on combined features $f = \{f^t, \hat{f}^{0\sim t-1}\}$.
			\STATE Train $K^{t}$ on $f = \{f^t, \hat{f}^{0\sim t-1}\}$ for next task.
			\ENDFOR 
		\end{algorithmic}
	\end{algorithm}

	\section{Experiments}
	
	\subsection{Implementation Details}
	We validate the reconstruction capability and storage efficiency of KRNet and compare it with the AE baseline. The proposed incremental learning method KRIL is evaluated on the two most commonly used benchmarks, i.e. CIFAR-100~\cite{Krizhevsky2009LearningML} and ImageNet-Subset~\cite{hou2019learning,rebuffi2017icarl}.

	\subsubsection{Datasets} The CIFAR-100 dataset consists of 100 categories. There are 500 images for training and 100 images for testing in each class. As in most previous works, we pad the images with 4 pixels and randomly crop a patch of $32 \times 32$. ImageNet-Subset contains 100 classes which are sampled from ImageNet~\cite{ILSVRC15} with an identical random seed (1993) following \cite{hou2019learning,rebuffi2017icarl,liu2020generative}. Within each class, there are about 1,300 training and 50 test images. We resize each image such that the short side has a size of 256, and randomly crop a patch of $224 \times 224$ during training. In testing stage, we replace the random crop with center crop.
	
	\subsubsection{Classification Networks}
	\label{cls_net}
	Following the conventional setting~\cite{hou2019learning,rebuffi2017icarl}, we use a modified 32-layer ResNet~\cite{he2016deep} for CIFAR-100 and the standard ResNet-18 for ImageNet-Subset. We split the network used in CIFAR-100 into the feature extractor ($F_1$) and the task learner ($F_2$) at the 11-th basic building block. That means $F_1$ and $F_2$ contain 23 and 9 layers respectively. The shape of features extracted by $F_1$ is $64 \times 8 \times 8$. For the standard ResNet-18, we split it at the 6-th basic building block, resulting in 13 layers for $F_1$ and 5 layers for $F_2$. $F_1$ of this network generates features of $256 \times 14 \times 14$.
	
	\subsubsection{KRNet}
	In our experiments, we set the group size $H$ and also the dimension of the static and dynamic vectors to 512. The output dimension of the two FC layers in the batched ID embedding module is kept the same as the input ($H$). In the feature decoder, we denote the output dimensions of the two FC layers as $d_0$ and $d_1$ respectively, the output channels of the convolutional layers before and after the deconvolution module as $c_0$ and $c_1$, and the stride of the deconvolution layer as $s_d$. The detailed setting of these hyper-parameters is shown in Table~\ref{setting-menet}. The kernel sizes of all convolutional layers and the deconvolution layer in KRNet are set to $3 \times 3$ and  $5\times5$ respectively. The number of groups in GN is 2 and the negative slope of Leaky-ReLU is $10^{-4}$. 
	For the sake of fairness, the decoders of KRNet and AE share the same architecture. And the encoder of AE is a symmetric mirrored network of the decoder. In the continual learning setting, we use two KRNets to recite the features of the base task and incremental tasks separately on account of the unbalanced number of classes between the base task ($t=0$) and incremental tasks ($t\ge 1$).
	
\begin{table}[!t]
	\caption{The hyper-parameter setting of KRNet in our experiments.}
	\label{setting-menet}
    \centering
		\begin{tabular}{c|c|c|c|c|c}
			\toprule
			Dataset         & $d_0$    &  $d_1$                    &  $c_0$ &  $c_1$  &  $s_d$\\ \midrule
			CIFAR-100       & 1024     & $512\times8\times8$   &  512   &  64     &  1\\
			ImageNet-Subset & 1536     & $1024\times7\times7$  &  1024  &  256    &  2\\ \bottomrule
		\end{tabular}
\end{table}

	\subsubsection{Training Details} 
	For CIFAR-100, we train the base classification task ($t=0$) for $160$ epochs with a batch size of $128$ and a weight decay of $5 \times 10^{-4}$ using the SGD~\cite{doi:10.1162/neco.1989.1.4.541} optimizer. The learning rate is initialized to $0.1$ and decayed by a scale of $0.1$ at epoch $80$ and $120$. When training the incremental tasks ($t\ge 1$), the learning rate of parameters in $\tilde{F}_2^t$ is multiplied by $0.05$, and the $\lambda$ in Eq.~\eqref{cls-loss} is set to $2$. The KRNet is trained using the Adam~\cite{Kingma2014Adam} optimizer with a batch size of 1000. The learning rate is fixed to $3\times 10^{-3}$ in the first $2 \times 10^{4}$ iterations and decayed linearly to $3\times 10^{-6}$ in the rest $2 \times 10^{4}$ iterations. The $\gamma$ in Eq.~\eqref{menet-loss} is set to $10^{-3}$. It is worth noting that weight decaying needs to be disabled to maximize the fitting ability of KRNet. We normalize the features linearly to the range of $[0, 1]$ for each channel. Because the features are extracted after the ReLU layer, their values are quite sparse. The ratio of zeros is $21\%$ for CIFAR-100 and $63\%$ for ImageNet-Subset. The sparsity of features eases the recitation task.
	For ImageNet-Subset, when training the base task, we set the total number of epochs to $100$. Accordingly, the epochs of learning rate reduction are changed to $50$ and $75$. When training the KRNet, we set $\gamma=5\times10^{-4}$ and increase the total training iterations to $5\times10^4$. All other settings are kept the same as CIFAR-100.
	
	
\subsection{Results}
\label{sec:results}

\begin{figure}[!t]
\centering
	\subfloat[Original]{\includegraphics[width=0.31\linewidth]{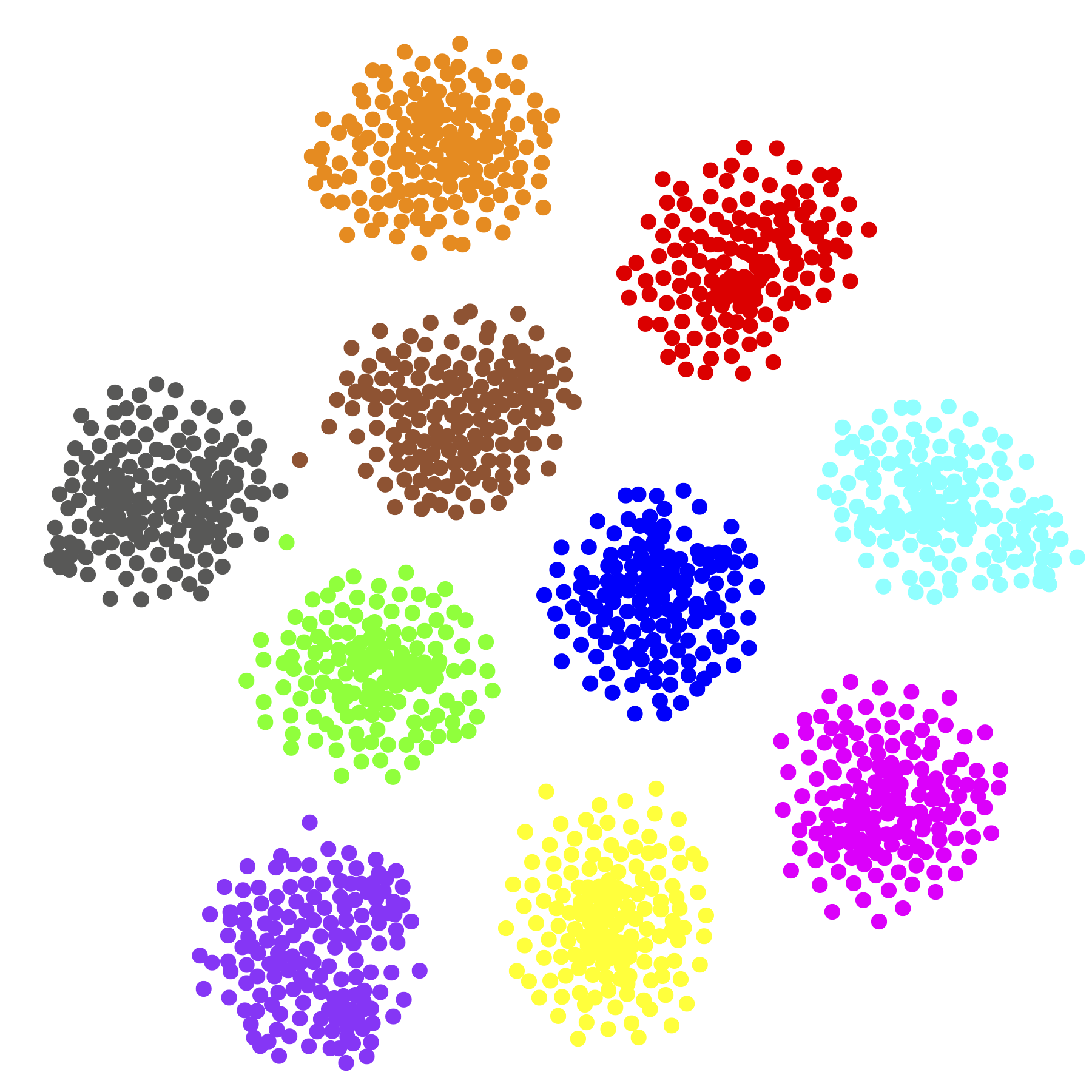}}
    \hfil
	\subfloat[KRNet]{\includegraphics[width=0.31\linewidth]{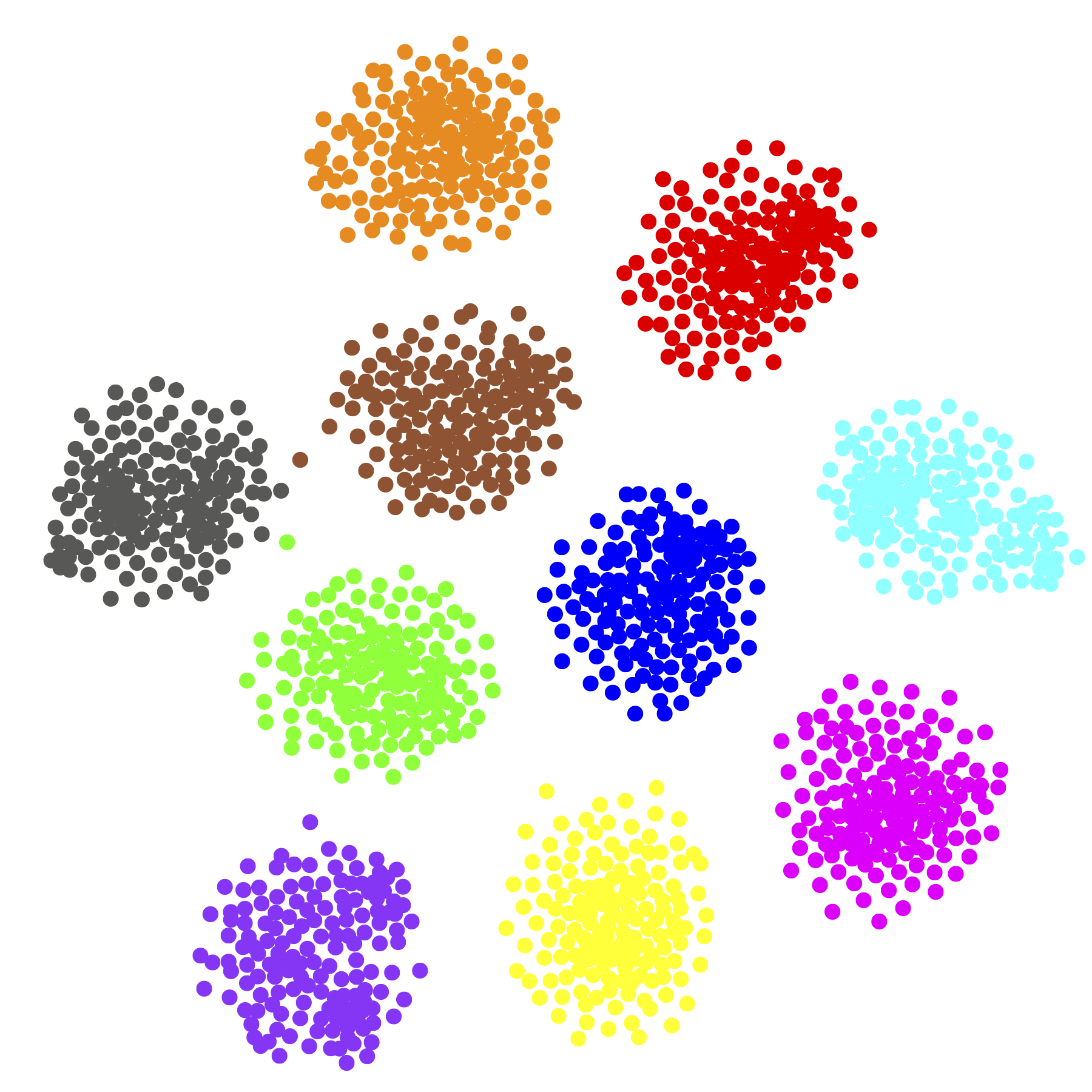}}
    \hfil
	\subfloat[AE]{\includegraphics[width=0.31\linewidth]{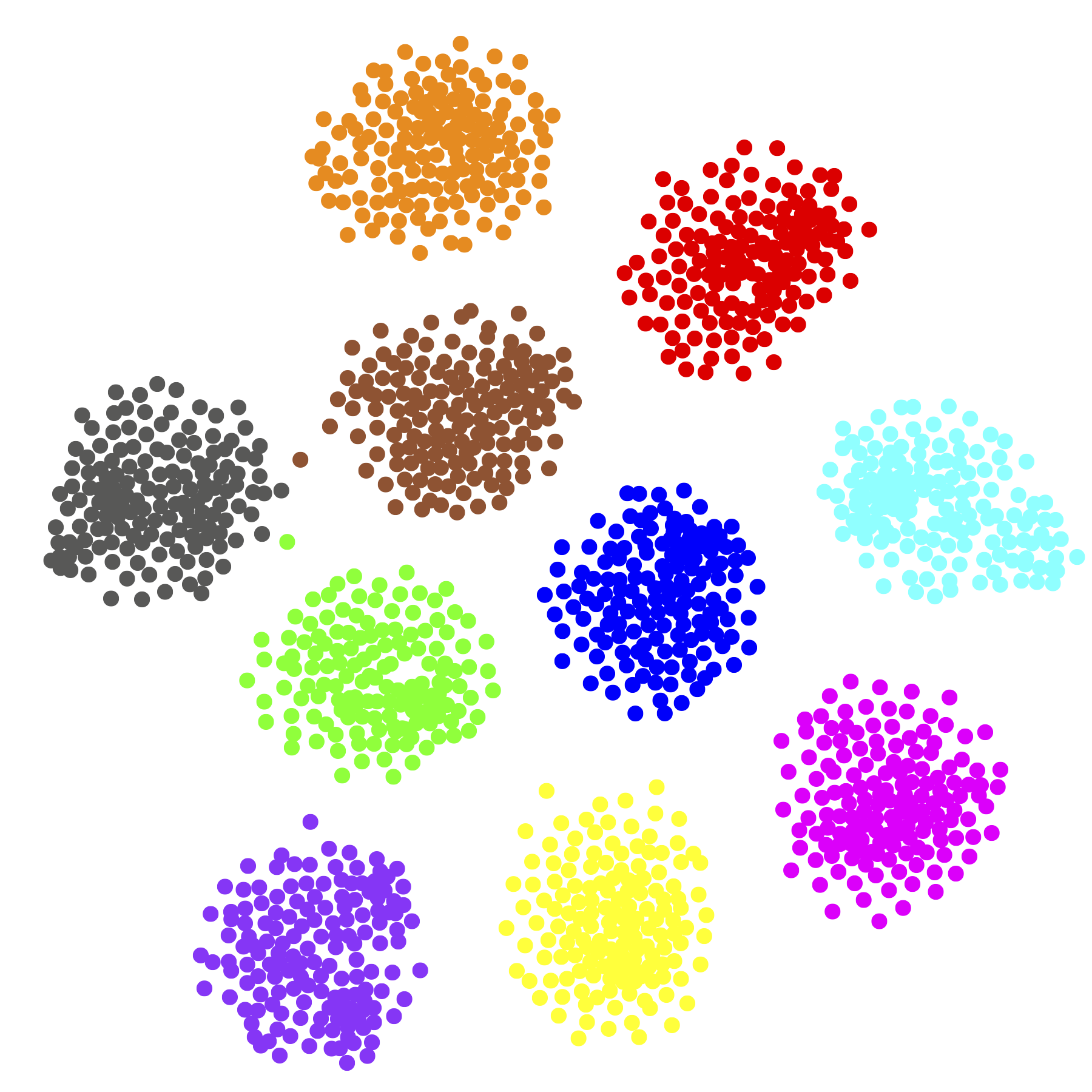}}
\caption{The t-SNE visualizations of the original features, KRNet replayed features and AE replayed features. For clarity we only show 10 random classes in ImageNet.}
\label{tsne}
\end{figure}

\begin{table}[!t]
    \caption{Comparison of different knowledge recording methods in terms of storage size under varying numbers of classes sampled from ImageNet. Feature storage measures the size of the raw features. AE storage and KRNet storage measure the size of latent codes required by AE and KRNet respectively.}
	\label{memory_cmp}
    \centering
		\begin{tabular}{r|r|c|c|c}
			\toprule
            \multirow{2}*{\# Classes} & \multirow{2}*{\# Samples} & Feature &AE & KRNet\\ 
			&  & Storage & Storage &Storage\\ 
			\midrule
		 	 50 & 64,817   & 12.12 GB  &253.19 MB  & 0.59 MB \\ 
			100 & 129,395  & 24.19 GB  &505.45 MB  & 1.17 MB \\ 
			150 & 194,217  & 36.30 GB  &758.66 MB  & 1.76 MB \\ 
			200 & 255,224  & 47.71 GB  &996.97 MB  & 2.34 MB \\ 
			250 & 319,811  & 59.78 GB  &  1.22 GB  & 2.93 MB \\ \bottomrule
		\end{tabular}
\end{table}

\begin{figure*}[!t]
    \centering
		\subfloat[CIFAR-100 (5 tasks)]{\includegraphics[width=0.24\linewidth]{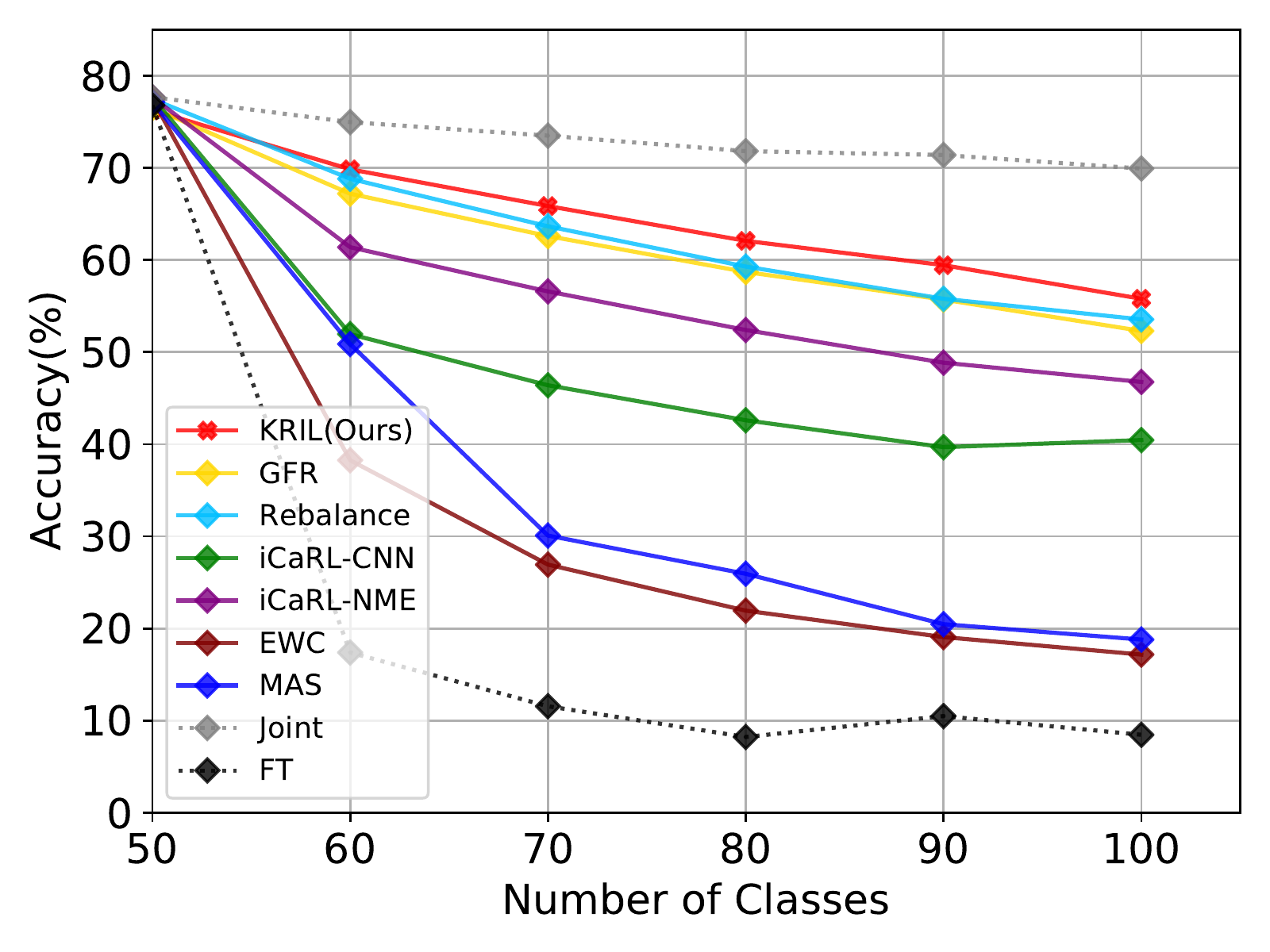}}
        \hfil
		\subfloat[CIFAR-100 (10 tasks)]{\includegraphics[width=0.24\linewidth]{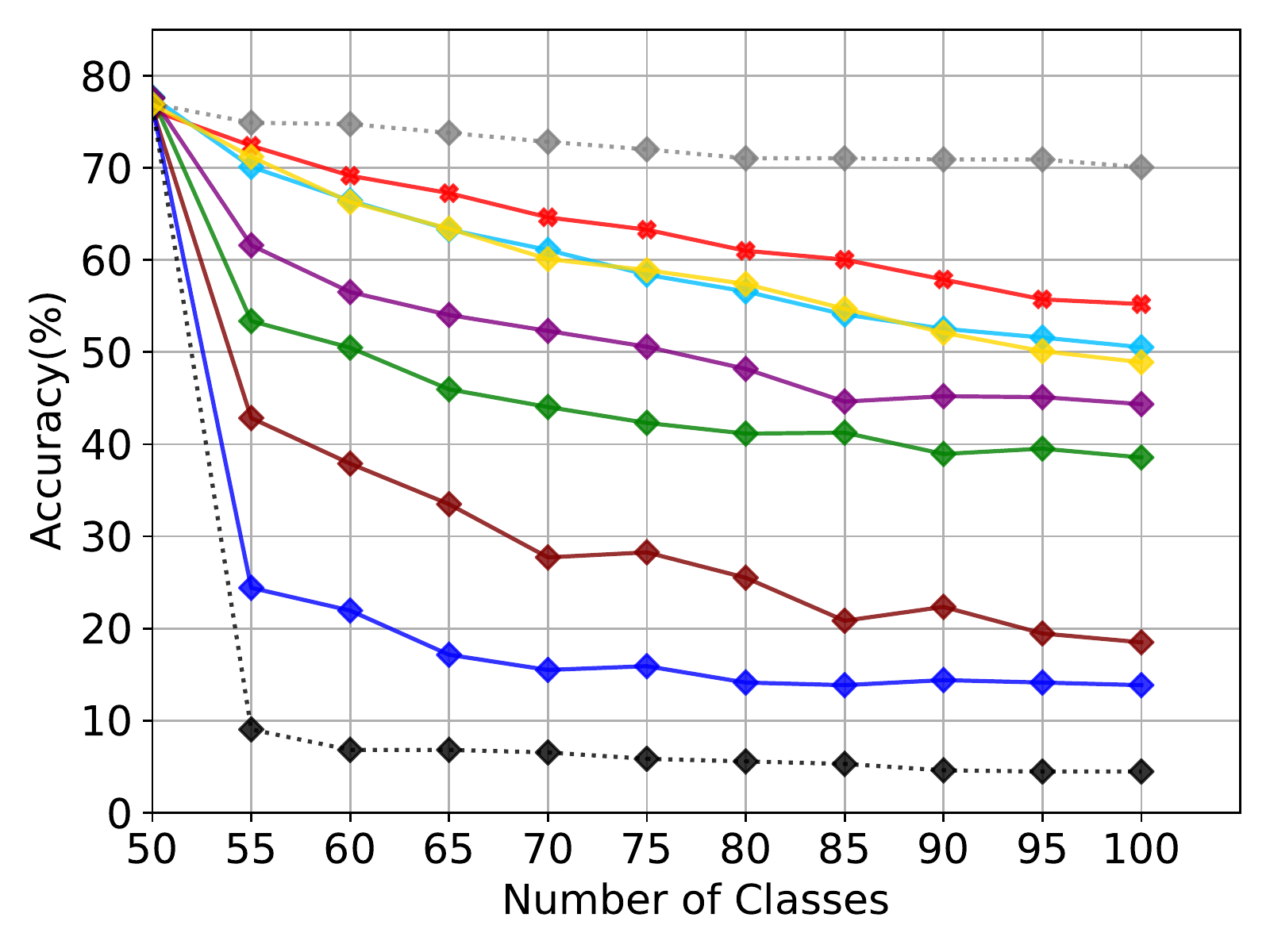}}
        \hfil
		\subfloat[ImageNet-Subset (5 tasks)]{\includegraphics[width=0.24\linewidth]{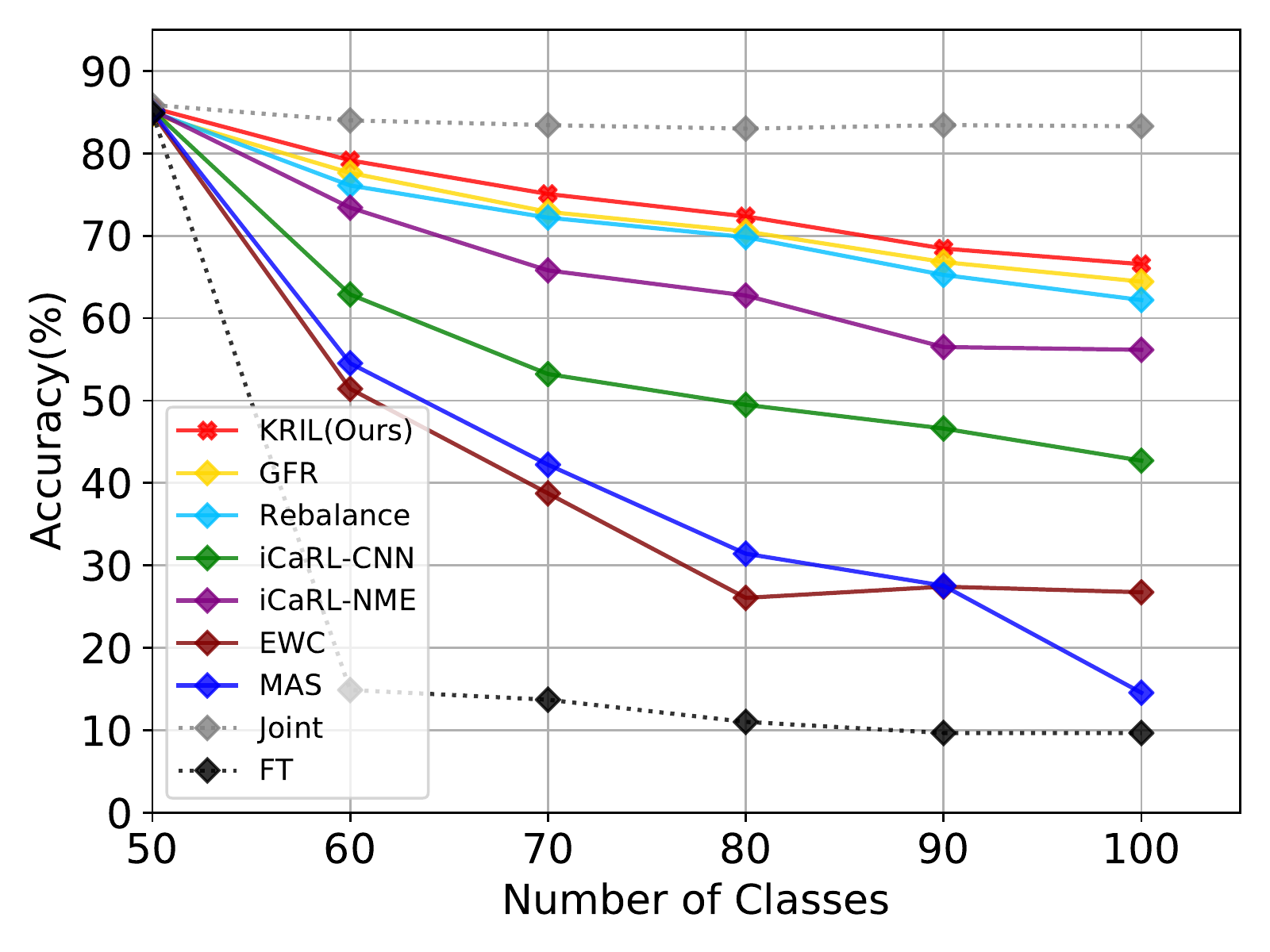}}
        \hfil
		\subfloat[ImageNet-Subset (10 tasks)]{\includegraphics[width=0.24\linewidth]{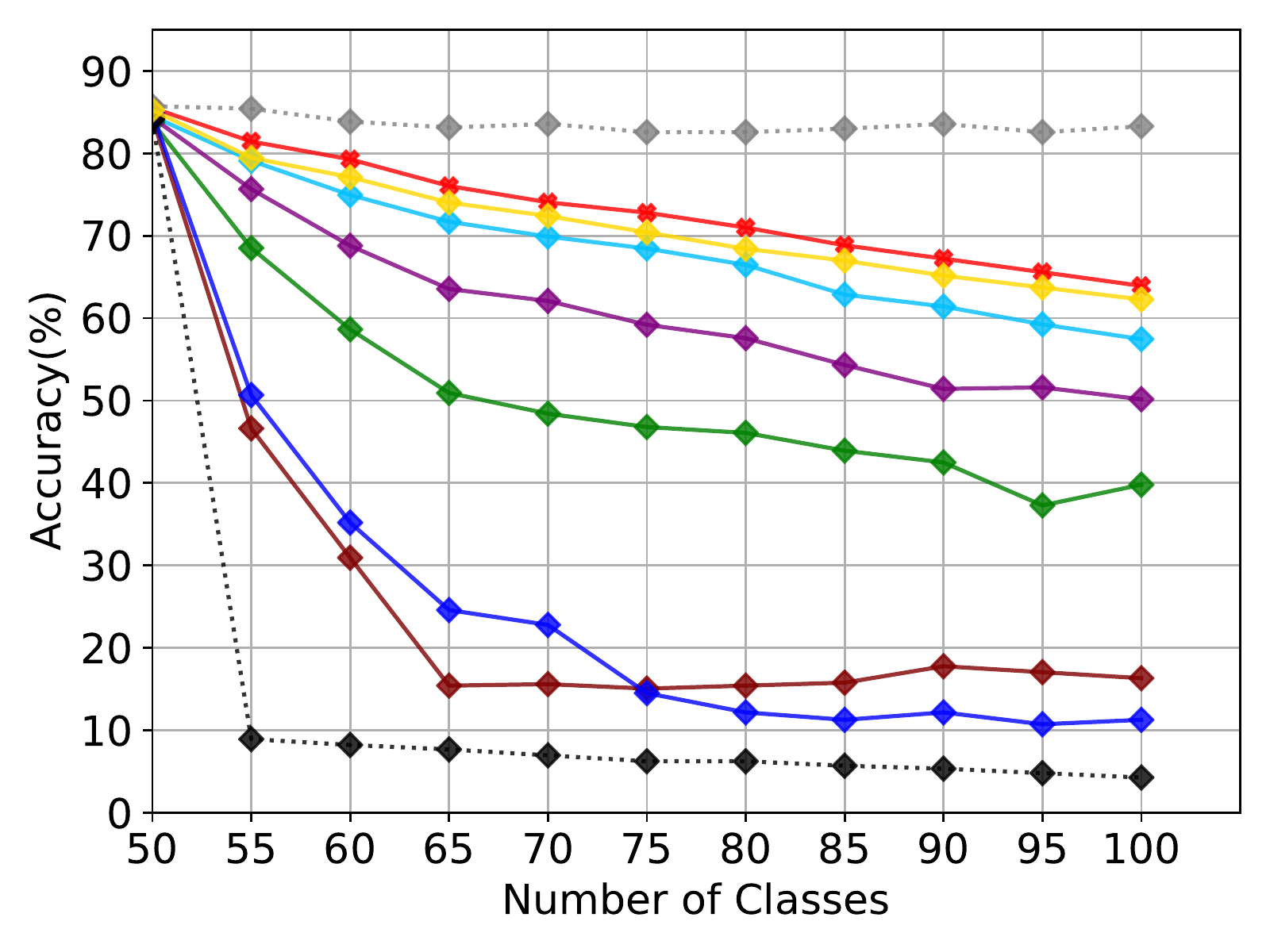}}
    \caption{Comparison with various existing methods in terms of classification accuracy on CIFAR-100 and ImageNet-Subset. Joint and FT refer to joint training (upper bound) and fine-tuning (lower bound) respectively.}
	\label{results}
\end{figure*}

	\subsubsection{Performance of KRNet}

    We validate the performance of KRNet in terms of reconstruction error and storage efficiency. When comparing it with the AE baseline, their training settings are kept as close as possible, e.g. with identical hyper-parameters and decoder architecture. From the experimental results on ImageNet, KRNet and AE show similar reconstruction capability. When recording about 64k features, the mean squared errors between the generated features and the original features are $4.63 \times 10^{-4}$ for KRNet and $6.65 \times 10^{-4}$ for AE. The t-SNE~\cite{maaten2008visualizing} visualizations of the original features and the features replayed by KRNet and AE are given in Fig.~\ref{tsne}. Both KRNet and AE can reconstruct the original features very well.
    
    The storage costs of the network weights of AE (without the encoder) and KRNet are similar, i.e. 325.8MB for AE and 327.9MB for KRNet. However, in terms of the size of latent codes, KRNet is significantly more efficient than AE. Table~\ref{memory_cmp} lists the size of latent codes required by the two methods. Compared with AE, KRNet requires $400\times$ less storage cost. Even for a large-scale dataset of 319k samples, the latent codes of KRNet only occupy 2.93MB. It is worth noting that the model size is fixed and negligible compared with the size of original features. The overall compression ratio increases with the scale of data, e.g. $36\times$ for 64k samples and $180\times$ for 319k samples. The extremely high compression ratio makes it highly competitive in knowledge replay applications.

	\subsubsection{Performance of KRIL}
	For both CIFAR-100 and ImageNet-Subset, we use half of the categories as the base task ($t=0$) and split the remaining classes evenly into $T$ (e.g. 5 and 10) tasks. We compare the KRIL framework with existing state-of-the-arts. GFR~\cite{liu2020generative} is the most relevant to our method, and is adopted as our baseline. Rebalance~\cite{hou2019learning} is a representative exemplar-based method. iCaRL~\cite{rebuffi2017icarl} with CNN prediction (iCaRL-CNN) and nearest-mean-of-exemplars classifier (iCaRL-NME) are compared. EWC~\cite{kirkpatrick2017overcoming} and MAS~\cite{aljundi2018memory} are two regularization-based methods without using any exemplars. Besides, we report the upper bound performance achieved by joint training and the lower bound by fine-tuning.
	
	The classification accuracy curves of different methods are shown in Fig.~\ref{results}. We can see the performances of rehearsal-based methods are much better than regularization-based methods. Our method outperforms all of the methods being compared in both 5-task and 10-task settings. In particular, compared with GFR, the accuracy of the final model over the full 100 classes gets improved by 6.3\% (from 48.9\% to 55.21\%) in the 10-task setting. Compared with Rebalance, we also achieve superior accuracy. And without storage of any exemplars, the risk of privacy leakage can be greatly mitigated. The results on ImageNet-Subset are consistent with those on CIFAR-100.

\subsection{Ablation Study}
	\subsubsection{Upper Bound Performance of KRIL}
\begin{figure}[!t]
    \centering
		\subfloat[CIFAR-100 (10 tasks)]{\includegraphics[width=0.48\linewidth]{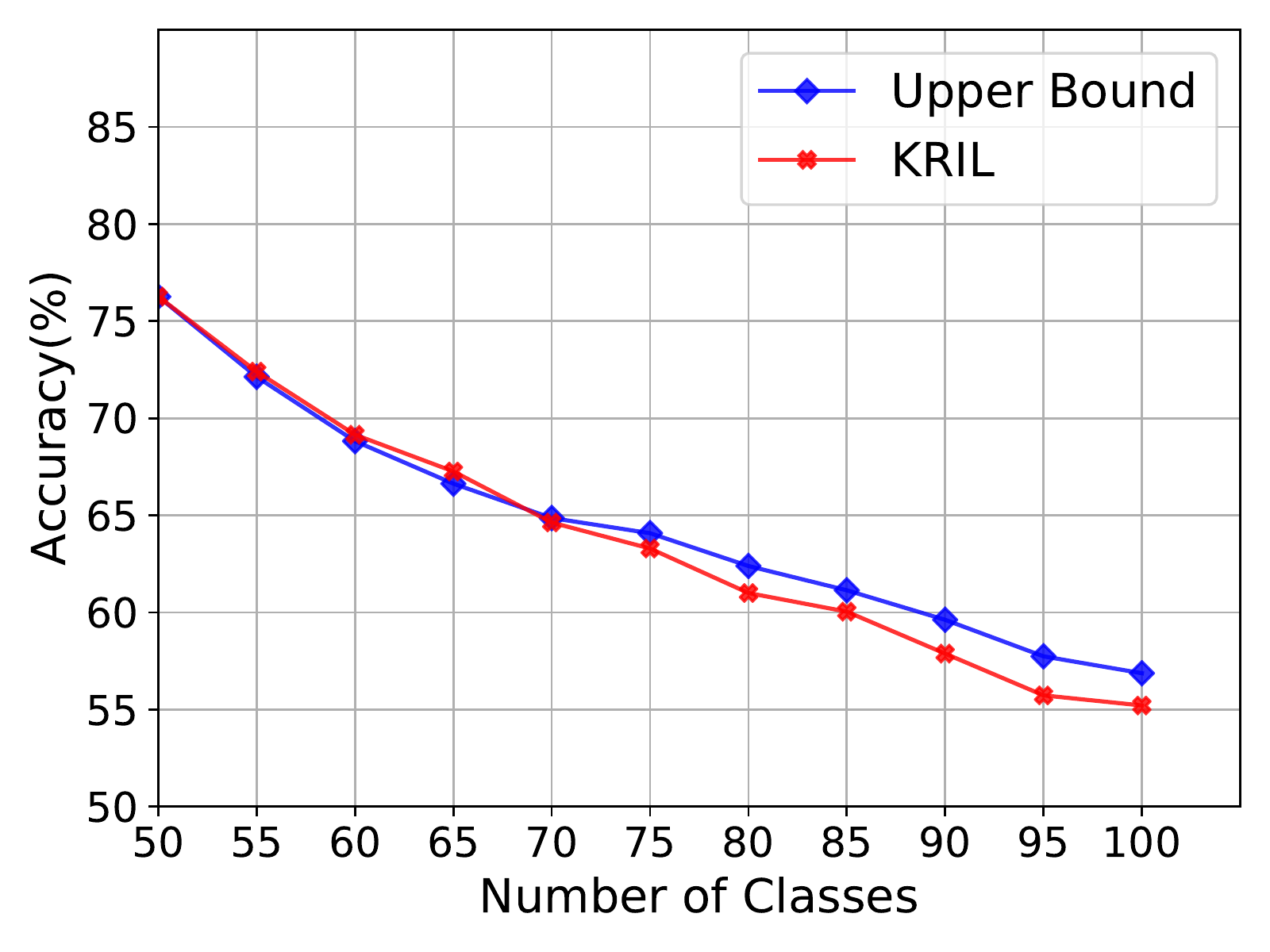}}
        \hfil
		\subfloat[ImageNet-Subset (10 tasks)]{\includegraphics[width=0.48\linewidth]{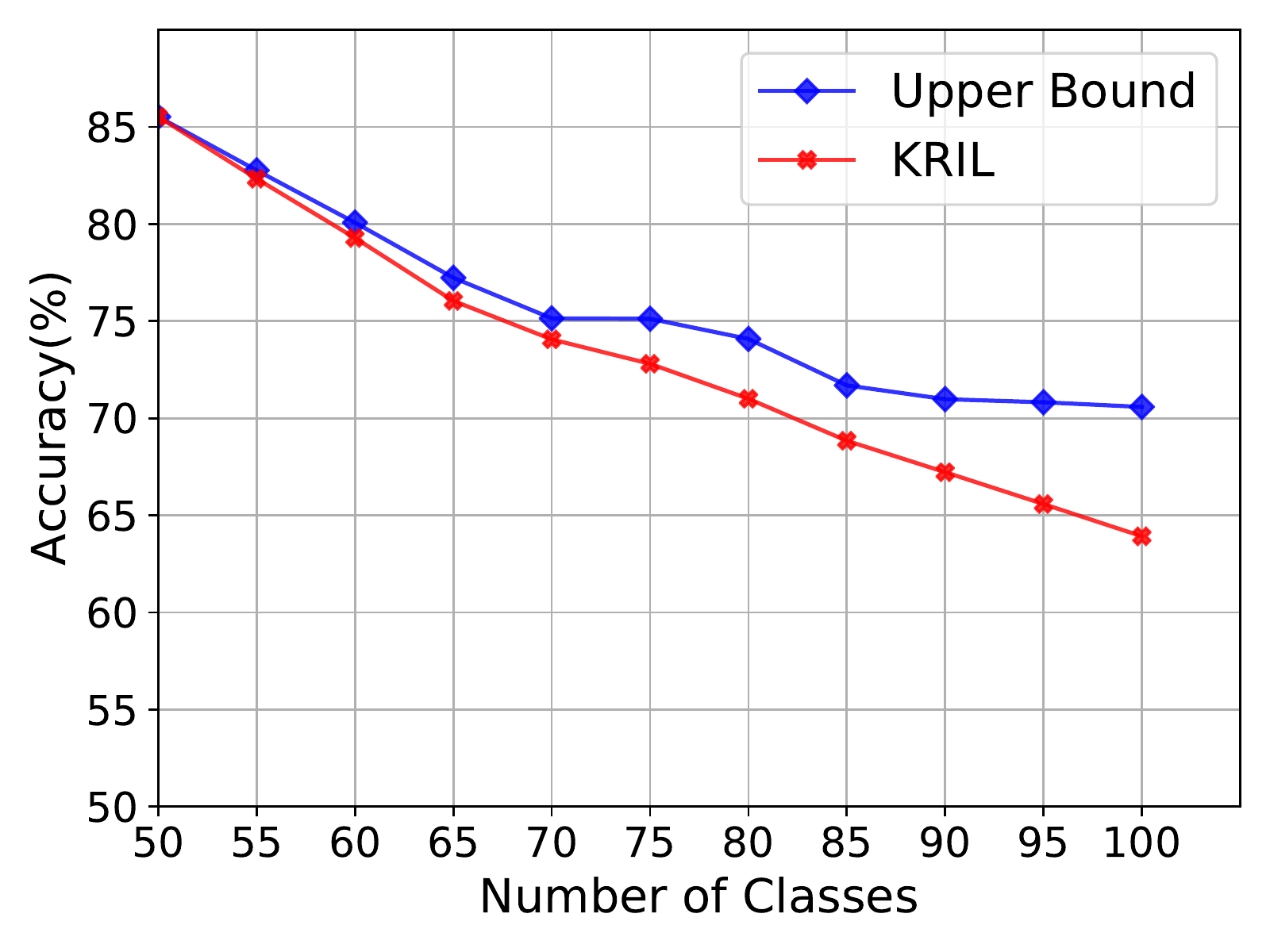}}
	\caption{The upper bound performance of KRIL achieved by using the original features rather than replayed features.}
	\label{upper_bound}
\end{figure}

	To investigate the feature recording ability of KRNet in the context of incremental learning, we measure the upper bound performance of KRIL by replacing the recited fake features with real features. As shown in Fig.~\ref{upper_bound}, the performance of KRIL is very close to the upper bound on CIFAR-100 (55.21\% vs. 56.87\%). The performance gap is relatively large on ImageNet-Subset (63.92\% vs. 70.58\%). This is because the dimension of features extracted from CIFAR-100 is much smaller than ImageNet-Subset ($64 \times 8 \times 8$ vs. $256 \times 14 \times 14$). It is much easier to recite features with a smaller dimension for KRNet. This observation indicates that there is still room for improvement in reciting high-dimensional features using a network, which is left as future work.

	\subsubsection{Single KRNet vs. Double KRNet}
\begin{figure}[!t]
    \centering
		\subfloat[CIFAR-100 (5 tasks)]{\includegraphics[width=0.48\linewidth]{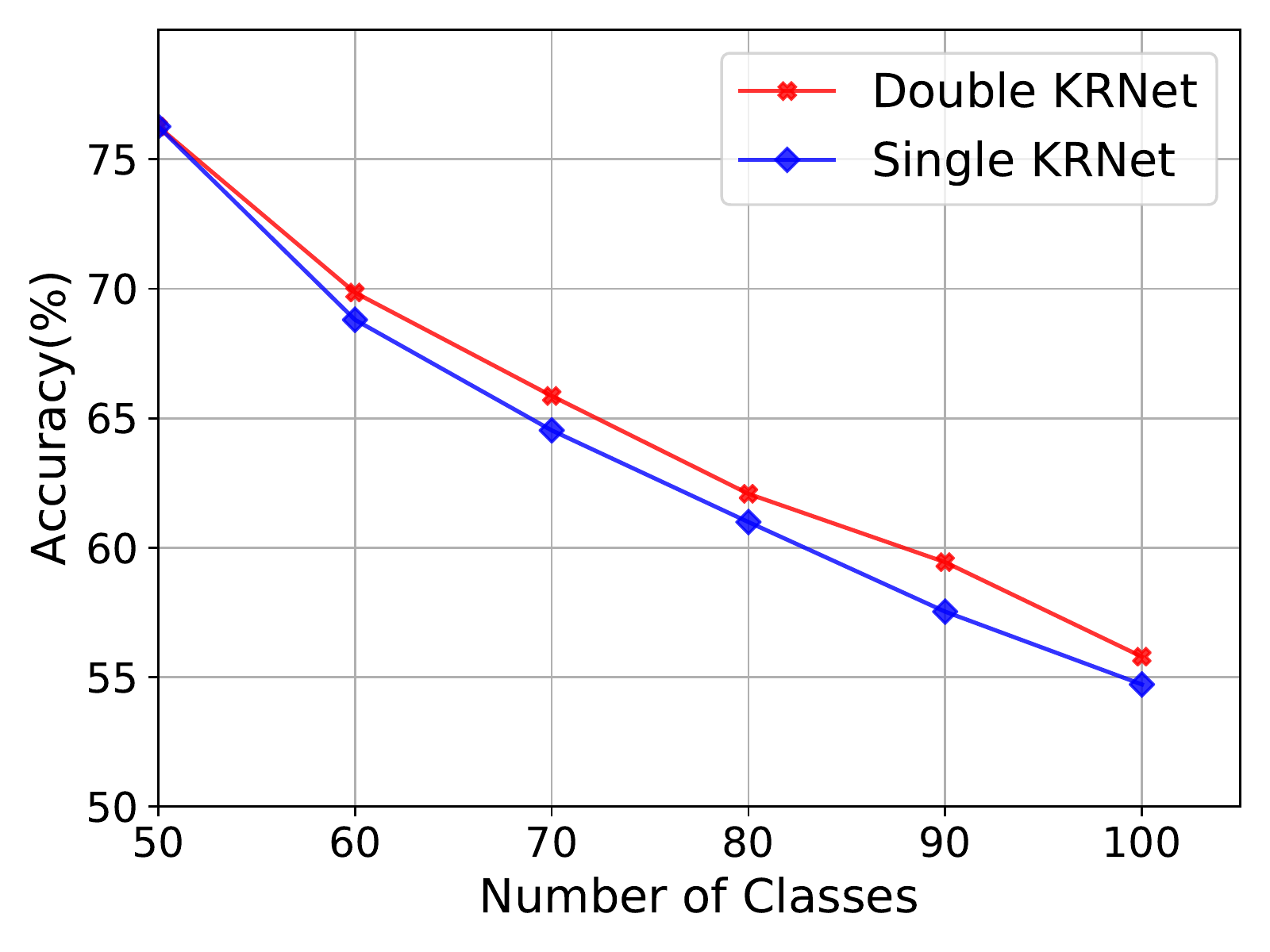}}
        \hfil
		\subfloat[CIFAR-100 (10 tasks)]{\includegraphics[width=0.48\linewidth]{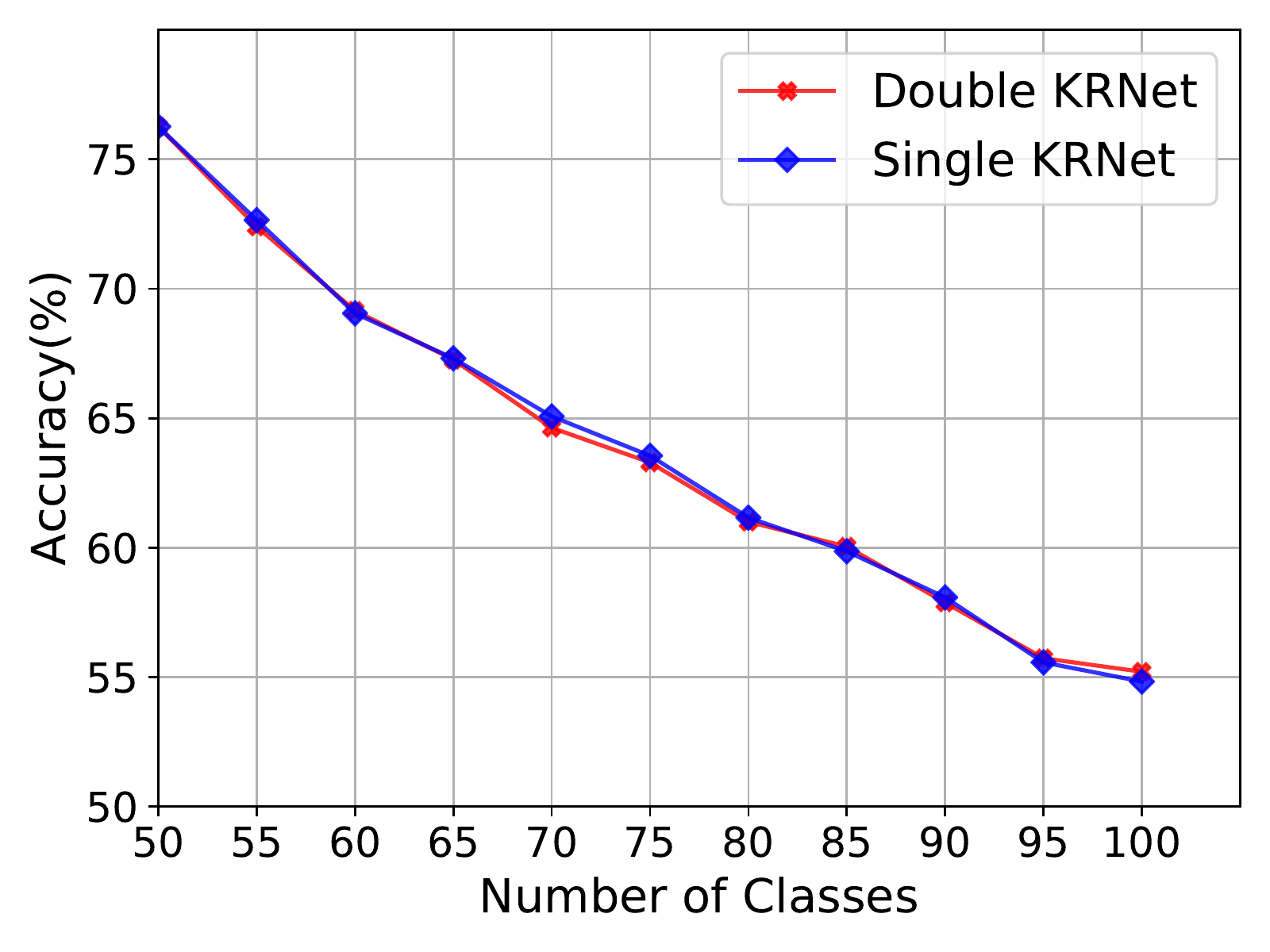}}
	\caption{Performance comparison between the single KRNet and double KRNet settings.}
	\label{single_vs_double}
\end{figure}

	In our KRIL, we use two KRNets to recite the features of the base task and the rest tasks separately. Alternatively we may also use a single KRNet to recite the features of all tasks. A comparison of performance is shown in Fig.~\ref{single_vs_double}. Single KRNet achieves very close performance to the double KRNet setting. In the 10-task setting, the performance gap is about $1\%$. Another advantage of using two KRNets is that the training time can be significantly reduced, as the initial task occupies half of the entire dataset.
	
	\section{Conclusion}
	
    With the widespread applications of deep learning, the knowledge represented by the intermediate feature maps rather than the raw images deserves more attention. How to efficiently encode the knowledge is critical to applications like continual learning. The large number of feature channels, varying floating point values covering a high dynamic range, and the sparsity of the features impose new challenges and opportunities. We propose a novel KRNet which takes advantage of the overfitting capability of DNNs to precisely encode the knowledge. Its superiority over the autoencoder baseline in terms of storage efficiency is well validated. Moreover, we show its application and present a novel incremental learning method. By replaying the knowledge of previous data encoded by KRNet, we achieve state-of-the-art performance on two incremental learning benchmarks. Our future work includes 1) improvement of KRNet in reciting large feature maps, and 2) more applications of KRNet such as continuous domain adaptation and arbitrary data compression.



\bibliographystyle{IEEEtran}
\bibliography{main}

\clearpage
\appendix

\subsection{Additional Ablation Study}

\subsubsection{Impact of the Auxiliary Loss in KRIL}
To validate the impact of the auxiliary loss $\mathcal{L}_{aux}$ in Eq.~(2), we evaluate the incremental learning performance when it is disabled. The performance comparison on the CIFAR-100 dataset is shown in Table~\ref{kril_aux_loss}. We can see that the accuracies between them are almost the same in the first eight incremental tasks. The importance of $\mathcal{L}_{aux}$ emerges in the last two tasks, where we observe an accuracy gain of $0.8\%$ and $1.4\%$. Since KRNet is trained recursively in a self-taught manner, the minor deviation of feature recitation in each task is accumulated. By minimizing the discrepancy between the feature predictions of current and last models, the erroneous replayed features are more tolerable. 
\begin{table}[h]
	\caption{Impact of the auxiliary loss $\mathcal{L}_{aux}$ to training of KRIL on CIFAR-100.}
	\label{kril_aux_loss}
    \centering
	\begin{tabular}{c|cc|c|cc}
		\toprule
		Task & w/ $\mathcal{L}_{aux}$ & w/o $\mathcal{L}_{aux}$ & Task & w/ $\mathcal{L}_{aux}$ & w/o $\mathcal{L}_{aux}$ \\ \midrule
		1 & 72.40 & 72.31  &  6   & 61.00   &  61.39\\ 
		2 & 69.15 & 69.08  &  7   & 60.04   &  59.88\\ 
		3 & 67.26 & 67.10  &  8   & 57.88   &  57.74\\ 
		4 & 64.63 & 64.41  &  9   & 55.73   &  54.95\\ 
		5 & 63.29 & 63.15  &  10  & 55.21   &  53.80\\ \bottomrule
	\end{tabular}
\end{table}

\subsubsection{Impact of the Auxiliary Loss in KRNet}
Similarly, we also investigate the impact of the auxiliary loss $\mathcal{L}_{kr2}$ in Eq.~(3) to training of KRNet. As shown in Table~\ref{krnet_aux_loss}, by using $\mathcal{L}_{kr2}$ we obtain little performance gain. However, $\mathcal{L}_{kr2}$ makes the training more stable. Considering the complexity and non-linearity of neural networks, a slight modification to the input may result in a large change in the output~\cite{su2019one}. In other words, the same numerical difference in $\hat{f}$ may have different levels of sensitivity from the perspective of $\tilde{F}_2^{t}$. $\mathcal{L}_{kr2}$ can help KRNet to predict more reasonable features and thus stabilize the subsequent incremental training.

\begin{table}[h]
	\caption{Impact of the auxiliary loss $\mathcal{L}_{kr2}$ to training of KRNet on CIFAR-100.}
	\label{krnet_aux_loss}
	\centering
	\begin{tabular}{c|cc|c|cc}
		\toprule
		Task & w/ $\mathcal{L}_{kr2}$ & w/o $\mathcal{L}_{kr2}$ & Task & w/ $\mathcal{L}_{kr2}$ & w/o $\mathcal{L}_{kr2}$ \\ \midrule
		1 & 72.40 & 72.24  &  6   & 61.00   &  61.14\\ 
		2 & 69.15 & 69.19  &  7   & 60.04   &  59.81\\ 
		3 & 67.26 & 67.25  &  8   & 57.88   &  57.54\\ 
		4 & 64.63 & 64.27  &  9   & 55.73   &  55.63\\ 
		5 & 63.29 & 63.16  &  10  & 55.21   &  54.64\\ \bottomrule
	\end{tabular}
\end{table}
   
\subsubsection{Number of Fixed Layers}
In our KRIL, we fix the first 23 layers of ResNet-32 used on the CIFAR-100 dataset. The remaining updatable part contains 4 basic building blocks and a fully-connected layer. Here we evaluate the performance when more layers are fixed. To rule out the effect brought by KRNet, we report the oracle incremental learning performance by using the original features. As compared in Table~\ref{impact_diffferent_layers}, as the task ID increases, the classification accuracy decays more quickly when more layers are fixed. The reason is that the more layers we fix, the fewer parameters could be fine-tuned in the task learner, which results in insufficient capacity for learning of new tasks. If we use features near the last layer of the backbone network for incremental learning, the feature extractor needs to be adapted for new data~\cite{belouadah2021a}. This is one of the important reasons why fine-tuning of the feature extractor with knowledge distillation loss is required in most feature replay-based methods.
\begin{table}[h]
	\caption{Oracle incremental learning performance with different number of layers being fixed.}
	\label{impact_diffferent_layers}
	\centering
	\begin{tabular}{c|cccc}
		\toprule
		\multirow{2}*{Task}& \multicolumn{4}{c}{\# Fixed Layers } \\  \cline{2-5} 
		& 23     & 25     & 27    & 29 \\ \midrule
		1 & 72.12  & 72.53  &72.09  & 71.76 \\ 
		2 & 68.82  & 68.65  &68.85  & 68.68 \\ 
		3 & 66.63  & 66.62  &66.74  & 66.28 \\ 
		4 & 64.86  & 64.24  &64.26  & 64.39 \\ 
		5 & 64.08  & 63.88  &63.56  & 63.67 \\ 
		6 & 62.39  & 62.50  &59.77  & 59.82 \\
		7 & 61.14  & 59.84  &58.59  & 58.35 \\
		8 & 59.62  & 58.30  &56.40  & 56.08 \\
		9 & 57.74  & 56.20  &54.73  & 54.21 \\
		10& 56.87  & 54.09  &53.22  & 52.88 \\ \bottomrule
	\end{tabular}
\end{table}

\subsection{Why to Replay Features Instead of Images}
In the proposed KRIL framework, we use KRNet to recite the features rather than images of previous tasks. The advantages are three-fold. 1) Samples of the same class tend to form clusters and be close to each other in the feature space. This is beneficial for the batched ID embedding which shares a dynamic vector in each group. 2) It is much easier for KRNet to recite features than images due to the sparsity and reduced dimension. 3) By sharing the low-level features, learning new tasks could be achieved by tweaking the deep layers only, which is more efficient than the incremental scheme using recited images. In addition, the proposed KRNet could be extended to other applications in which a large number of features are helpful, such as unsupervised domain adaptation~\cite{volpi2018adversarial} and anomaly detection~\cite{cohen2020sub}.

\end{document}